\documentclass[conference]{IEEEtran}
\IEEEoverridecommandlockouts
\usepackage{cite}
\usepackage{array}
\usepackage{amsmath,amssymb,amsfonts}
\usepackage{algorithmic}
\usepackage{graphicx}
\usepackage{textcomp}
\usepackage{xcolor}
\usepackage{url}
\usepackage{booktabs}
\def\BibTeX{{\rm B\kern-.05em{\sc i\kern-.025em b}\kern-.08em
    T\kern-.1667em\lower.7ex\hbox{E}\kern-.125emX}}
\begin{document}

\title{UniView: Enhancing Novel View Synthesis from a Single Image by Unifying Reference Features}

\author{Haowang Cui~~~~   Rui Chen~~~~   Jiaze Wang~~~~~   Tao Guo~~~~~   Zheng Qin\\
Tianjin University\\
{\tt\small\{haowangcui, ruichen, jiaze\_w, guotao\_stu, zhengqin58\}@tju.edu.cn}}

\maketitle

\begin{abstract}
The task of synthesizing novel views from a single image is highly ill-posed due to multiple explanations for unobserved areas. Most current methods tend to generate unseen regions from ambiguity priors and interpolation near input views, which often lead to severe distortions. To address this limitation, we propose a novel model dubbed UniView, which can leverage reference images from a similar object to provide strong prior information during view synthesis. More specifically, we construct a retrieval and augmentation system and employ a multimodal large language model (MLLM) to assist in selecting reference images that meet our requirements. Additionally, a plug-and-play adapter module with multi-level isolation layers is introduced to dynamically generate reference features for the target views. Moreover, in order to preserve the details of an original input image, we design a decoupled triple attention mechanism, which can effectively align and integrate multi-branch features into the synthesis process. Extensive experiments have demonstrated that our UniView significantly improves novel view synthesis performance and outperforms state-of-the-art methods on challenging datasets.
\end{abstract}


\section{Introduction}

\begin{figure}
    \centering
    \includegraphics[width=0.9\linewidth]{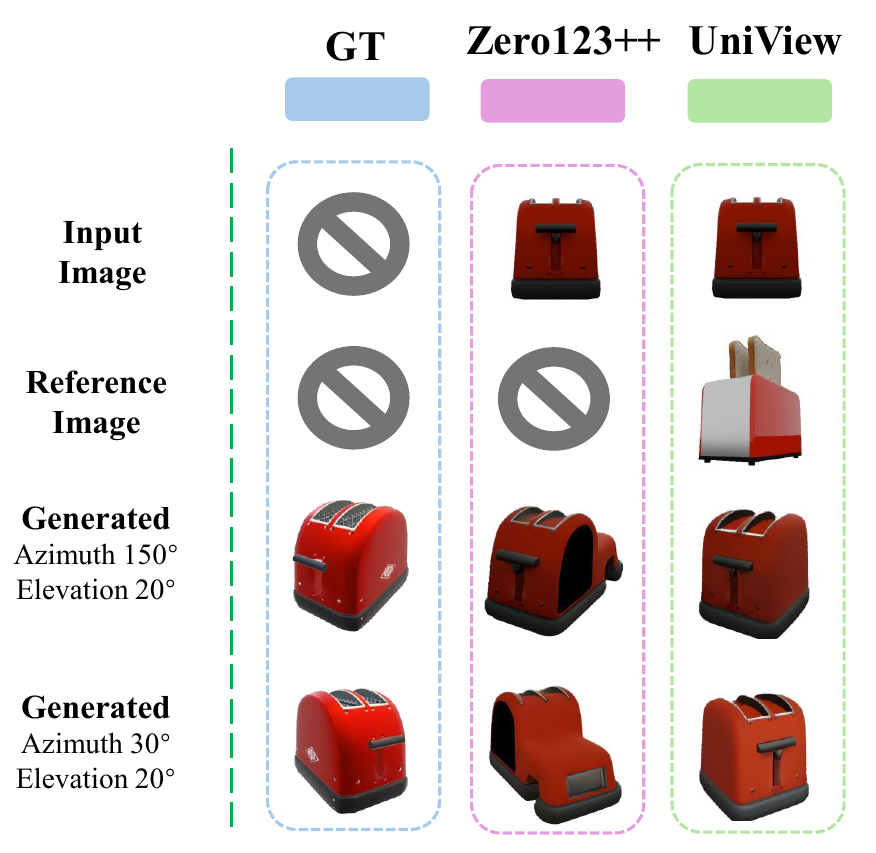}
    \vspace{-0.4cm}
    \caption{Motivation. Standard single-image NVS models (e.g., Zero123++) fail to synthesize occluded regions. UniView utilizes a reference image from a similar object to guide the synthesis, restoring correct geometry.}
    \label{fig:teaser}
    \vspace{-0.4cm}
\end{figure}

Novel View Synthesis (NVS) is fundamental to 3D generation and reconstruction\cite{One-2-3-45++, 3DGS, nerf}. While multi-view methods excel, acquiring dense views is impractical, making single-image NVS crucial. Recent diffusion-based frameworks\cite{zero123, NVSwith3Ddiffusion} treat this as image-to-image translation but remain highly ill-posed due to the lack of information in unobserved regions. Models often hallucinate appearances, leading to severe distortions (e.g., the artifact shown in Fig. \ref{fig:teaser}).


To address this, TOSS\cite{toss} introduced the use of text prompts as control conditions to provide global information, which could enhance the quality of back-side generation in single-image novel view synthesis. However, text is inherently limited in its capacity to accurately depict object features, and relying solely on text often results in insufficiently precise control.

"Good artists copy, great artists steal."  – Pablo Picasso's adage reveals an essential truth in artistic creation: faithful recreation stems from strategic appropriation. Inspired by Picasso's philosophy, we propose that "Good models generate, great models transplant." Acquiring multi-view images of a target object has been challenging in practical scenarios, while obtaining images of arbitrary objects from the same category is relatively straightforward. For instance, when only possessing a single image of a specific toaster, we can utilize complementary viewpoint images which are from different toaster within the same category as reference for novel view synthesis. Following this principle, we present UniView - a novel framework which is capable of transferring complementary visual information from similar reference objects (e.g., the back view of toaster B)  to guide novel view synthesis of target objects (e.g., toaster A). As demonstrated in Figure \ref{fig:teaser}, image-based references from complementary regions of analogous objects serve as a powerful and precise control signal, significantly improving the synthesis quality of unobserved regions in the original input view.

However, the inherent misalignment dilemma between reference images and target objects makes it challenging to inject reference information into the pre-trained multi-view diffusion model while avoiding misleading guidance for novel view synthesis. In some scenarios, reference image acquisition would be a hindrance. To address these issues, as depicted in Figure \ref{fig:figure3}, we design a complete system which leverages a multimodal LLM to retrieve the most appropriate reference image from the pre-constructed database. Also, UniView introduces an adaptive architecture that dynamically modulates the intensity of the reference signal, which ensures precise and effective control of reference images over the pre-trained single-image novel view synthesis model. The main contributions of this work are summarized as follows:

First, we design a  \textbf{Dynamic Reference Retrieval System} to automatically select optimal reference images. Second, we propose a \textbf{Meta-Adapter Module} that generates adaptive dynamic reference signals from input reference images to enhance novel view synthesis quality for target objects. Finally, we develop a \textbf{Decoupled Triple Attention Mechanism} that introduces global control information from reference images into the model while preserving the pre-trained model's inherent novel view synthesis capability.

\section{Related work}

\subsection{Diffusion models for 3D generation}
Building on the success of 2D diffusion models, recent works have extended them to 3D generation. A prominent stream utilizes Score Distillation Sampling (SDS)~\cite{dreamfusion} to distill 2D priors for 3D generation~\cite{text-to-3D-withCFD}. Alternatively, native 3D diffusion methods train directly on 3D representations, such as NeRF~\cite{diffrf}, 3D Gaussian Splatting~\cite{gvgen}, and Triplanes~\cite{rodin}. Specifically for novel view synthesis, diffusion models are leveraged to enforce multiview consistency~\cite{NVSwithDiffusionl} or fine-tune 2D priors on 3D data~\cite{zero123, zero123++}, enabling high-fidelity generation from sparse inputs.

\subsection{Novel view synthesis}
Traditional Novel View Synthesis (NVS) relies on explicit geometry reconstruction. The field was revolutionized by Neural Radiance Fields (NeRF)~\cite{nerf}, which model scenes as continuous volumetric fields. Recently, 3D Gaussian Splatting~\cite{3DGS} has emerged as a superior alternative, offering real-time rendering speeds. While these methods excel with dense views, single-image NVS remains challenging. Current state-of-the-art approaches have shifted from deterministic regression~\cite{synsin} to generative frameworks~\cite{zero123}, harnessing the hallucination capabilities of pre-trained diffusion models to synthesize plausible views from a single image.



\section{Methodology}
\begin{figure*}[!t]
    \centering
    \includegraphics[width=0.95\textwidth]{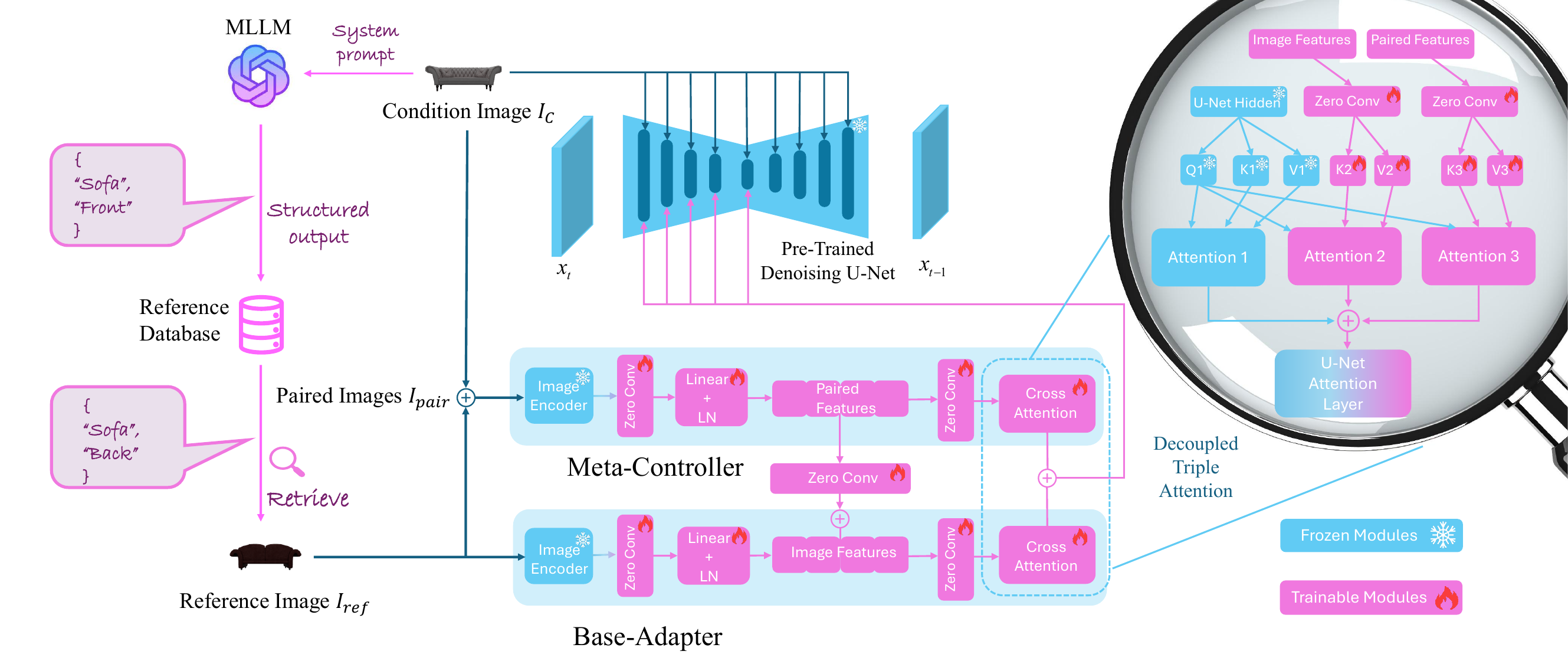}
    \caption{The architecture of UniView. The system leverages a multimodal large model to retrieve the optimal reference image from the database based on the input condition image. Then, an image pair composed of a condition image and a reference image is processed through a Meta-Adapter, which integrates a Base-Adapter and a Meta-Controller. Subsequently, the output is incorporated into a pre-trained multi-view diffusion model via a Decoupled Triple Attention mechanism. Zero convolution layers are strategically inserted between the Base-Adapter and Meta-Controller modules, as well as preceding the Decoupled Triple Attention mechanism, to ensure effective isolation.}
    \label{fig:figure3}
    \vspace{-0.5cm}
\end{figure*}

Our goal is to establish a novel view synthesis framework that simultaneously accepts both a condition image and a complementary-view reference image from the same object category as input. To provide our model with appropriate reference images, we construct a \textbf{Dynamic Reference Retrieval System} (Section \ref{sec:Dynamic Reference Retrieve}) which is inspired by the Retrieval-Augmented Generation (RAG)\cite{RAG} technique employed in large language models. This system utilizes a multimodal large language model (MLLM) to automatically select images from the database that match the input image, serving as reference images for subsequent processing stages.


To leverage the reference image for global guidance without interfering with the condition image, we propose a reference processing module that adaptively injects extracted features into a frozen multi-view diffusion backbone. While our framework is compatible with various architectures, we specifically employ Zero123++ \cite{zero123++} as the base model for our experiments.

We denote the condition image as \(I_{c}\) and the reference image as \(I_{ref}\). The training objective of our diffusion model \(\mathrm{\epsilon}_{\mathrm{\theta}}\) can be then formulated as:
\begin{equation}
    \mathcal{L} \,\,=\mathbb{E} _{t,\epsilon \sim \mathcal{N} \left( 0,1 \right)}\,\left[ \left\| \epsilon -\epsilon _{\theta}\left( x_t,t,I_c,I_{ref} \right) \right\| ^2 \right] 
\end{equation}
where \(t\) is the time step, \(x_{t}\) is denoised feature at time \(t\) and \(\mathrm{\epsilon}\) is the ground-truth noise.

To maximize the utilization of the pre-trained prior, only the additional conditioner used for processing the reference image is trainable, while the base multi-view diffusion model remains frozen.

Two dominant approaches exist for adding image conditions to diffusion models: the Adapter\cite{t2i-adapter} methodology and the ControlNet\cite{controlnet} framework. However, directly applying these methods to our task causes over-alignment with the reference image. This severely degrades the novel view synthesis capability of the conditional generation framework. To address this challenge, we implement: \textbf{Meta-Adapter Module} (Section \ref{sec:Meta-Adapter}) with multilevel isolation for adaptive dynamic control of the conditioning strength and \textbf{Decoupled Triple Attention Mechanism}  (Section \ref{sec:Decoupled-Triple-Attention}) for effective feature injection.
The detailed architecture of our model is illustrated in Figure \ref{fig:figure3}.

\subsection{Dynamic Reference Retrieval System}
\label{sec:Dynamic Reference Retrieve}
Our method allows users to manually provide a reference image \(I_{ref}\) and a condition image \(I_{c}\) as inputs. However, in some cases, obtaining a reference image with a complementary perspective of the same category as the condition image might be difficult. Therefore, we construct a Dynamic Reference Retrieval System to automatically select appropriate images from a database as reference images when users can not provide them.

We constructed a database comprising 20,000 images from 100 object categories. Each category contained 200 instances captured from four canonical viewpoints (front, back, left, right), with category and viewpoint annotations. As illustrated in Figure \ref{fig:figure3}, when the system receives a condition image, we employ the multimodal large language model GPT-4o\cite{gpt4o} to assist in selecting the reference image. Specifically, we predefined a system prompt that instructs the MLLM to infer the category and approximate viewpoint of the condition image, then return a structured JSON output. If the object does not belong to any of the 100 predefined categories, our system prompt guides the MLLM to provide the closest matching category label from the available 100 classes as a substitute. After receiving the MLLM’s response, we parse the formatted output, first retrieving objects of the same category as the condition image, and then selecting the corresponding complementary viewpoint image as the reference image.

\subsection{Meta-Adapter Module}
\label{sec:Meta-Adapter}
We adopt a lightweight adapter architecture rather than ControlNet\cite{controlnet} to inject conditions into the frozen model, minimizing trainable parameters. However, in our setting, the reference image is from the same category as the condition image but represents a different instance, which typically does not maintain strict alignment with the condition image. Original adapters fail in this scenario by enforcing strict alignment, which introduces conflicting features from the unaligned reference. To this end, we propose Meta-Adapter, as shown in Figure \ref{fig:figure3}, which could dynamically adjust the control strength of the reference image.

The Meta-Adapter comprises two components: a Base-Adapter and an additional Meta-Controller. The Base-Adapter, denoted as \(\mathcal{F} _{\varTheta}^{base}\left( \cdot \right) \), is composed of a frozen image encoder, trainable zero convolution layers and trainable dimension-matching layers including linear and linear normalization layers for feature dimension alignment.

The Meta-Controller, denoted as \(\mathcal{F} _{\varTheta'}^{meta}\left( \cdot \right) \), shares an identical architecture with the Base-Adapter while maintaining different trainable parameters \(\varTheta'\). It takes a set of paired  \(I_{c}\) and \(I_{ref}\) as input. After passing through the image encoder, the paired features are concatenated into a single feature before the linear and layer normalization layers. As the Meta-Controller and Base-Adapter are jointly trained, the dynamic control signals generated by the Meta-Controller are inherently adaptive. The meta control signals generated by the Meta-Controller are injected into the pre-trained diffusion model through two distinct mechanisms. On the one hand, the control signals are first processed through a zero convolution layer and subsequently combined with the output of the Base-Adapter via summation before going through the decoupled triple attention. The output of this pathway can be formally denoted as: \(y_{meta1}\,\,=\,\,\mathcal{F} _{\varTheta'}^{meta}\left( I_{pair} \right) \) where \(I_{pair}\) is paired \(I_{c}\) and \(I_{ref}\). Then, the complete output of the Base-Adapter can be expressed as:
\vspace{-0.2cm} 
\begin{equation}
y_{base}\,\,=\,\,\mathcal{F} _{\varTheta}^{base}\left( I_{ref}, y_{meta1} \right) 
\end{equation}
On the other hand, the meta signals \(y_{meta2}\) generated by the Meta-Controller are directly injected into the pre-trained diffusion model via decoupled triple attention. Totally, the adaptive dynamic control signals \(y_{control}\) produced by the Meta-Adapter can be formulated as:
\vspace{-0.2cm} 
\begin{equation}
y_{control}\,\,=\,\,\mathrm{Cross} \mathrm{Attention}\left( y_{base}, y_{meta2} \right) 
\label{equation 3}
\end{equation}

The Meta-Controller learns an implicit gating mechanism end-to-end. By minimizing training loss on conflicting pairs, it adaptively modulates \(y_{meta1}\) and \(y_{meta2}\) to suppress misleading guidance from misaligned references while preserving useful priors.

Zero convolution layer is a convolution layer with all parameters initialized to zero, which can isolate newly introduced control conditions. In UniView, we introduce zero convolution layers at multiple locations. As shown in Figure \ref{fig:figure3}, we position trainable zero convolution layers at three essential locations: first after the image encoders of both Base-Adapter and Meta-Controller, second at their mutual interconnection point, and finally immediately before the decoupled cross-attention module. We denote zero convolution layers as \(\mathcal{Z} \left( \cdot \right) \), the signal outputted by Meta-Controller before summed with Base-Adapter can be formally expressed as:
\begin{equation}
y_{meta1}' =\,\,\mathcal{F} _{\varTheta'}^{meta}\left( \mathcal{Z} \left( I_{pair} \right) \right) 
\end{equation}
and the complete output of Base-Adapter with isolation can be expressed as:
\begin{equation}
y_{base}' =\,\,\mathcal{F} _{\varTheta}^{base}\left( \mathcal{Z} \left( I_c, y_{meta1}' \right) \right) 
\end{equation}

Totally, with multilevel isolation, the adaptive dynamic control signals \(y_{control}'\) produced by the Meta-Adapter can be formulated as:
\vspace{-0.2cm}
\begin{equation}
y_{control}'\,\,=\,\,\mathrm{CrossAttention}\left( \mathcal{Z} \left( y_{base}',y_{meta2}' \right) \right) 
\end{equation}
\vspace{-0.5cm}

These zero convolution layers ensure that during the initial training phase, the parameters of the Meta-Controller exert no influence on the Base-Adapter, while the entire Meta-Adapter's parameters also remain neutral to the original pre-trained diffusion model. The mechanism shields the diffusion backbone from initialization-induced interference, preserving the base model's integrity.

\subsection{Decoupled Triple Attention Mechanism}
\label{sec:Decoupled-Triple-Attention}

Our proposed method introduces a novel decoupled triple attention mechanism to effectively integrate reference information and control signals into a pre-trained multi-view diffusion model. This tripartite paradigm processes three distinct inputs through parallel cross-attention pathways: reference information, control signals, and original image features. By summing the outputs of these pathways, our model achieves a fine-grained injection of external conditions, enabling it to generate high-quality, multi-view outputs that are both consistent with the reference and guided by precise control signals.

We inject control signals into the down blocks and middle block of U‑Net to influence global structure. For the down blocks and the middle block in base diffusion model, given the hidden feature \(f_{base}\) before attention computation, the attention feature Z can be defined by the following equation: 
\begin{equation}
\mathrm{Z} =\,\,\mathrm{Attention}\left( \mathrm{Q},\mathrm{K},\mathrm{V} \right) \,\,=\,\,\mathrm{Softmax} \left( \frac{\mathrm{QK}^{\top}}{\sqrt{\mathrm{d}}} \right) \mathrm{V}
\end{equation}
where \(\mathrm{Q}=\,\,f_{base}\mathrm{W}_q\), \(\mathrm{K}=\,\,f_{base}\mathrm{W}_k\), \(\mathrm{V}=\,\,f_{base}\mathrm{W}_v\) are the query, key, and values matrices of the attention operation. \(\mathrm{W}_q\),  \(\mathrm{W}_k\),  \(\mathrm{W}_v\) are the weight matrices of the linear projection layers.  \(f_{base}\), \(\mathrm{W}_q\), \(\mathrm{W}_k\) and  \(\mathrm{W}_v\) are from the U-Net of base multi-view diffusion model and are all frozen.

Given the output hidden feature from the Base-Adapter \( y_{base}\), cross-attention \(\mathrm{Z}'\) can be computed as:
\begin{equation}
\mathrm{Z}'=\,\,\mathrm{Attention}\left( \mathrm{Q},\mathrm{K}',\mathrm{V}' \right) \,\,=\,\,\mathrm{Softmax} \left( \frac{\mathrm{QK}'^{\top}}{\sqrt{\mathrm{d}}} \right) \mathrm{V}'
\end{equation}
where \(\mathrm{Q}=\,\,f_{base}\mathrm{W}_q\), \(\mathrm{K}' =\,\,y_{base}\mathrm{W}'_k\), \(\mathrm{V}' =\,\,y_{base}\mathrm{W}'_v\) are the key, and values matrices from the Base-Adapter, \(\mathrm{W}'_k\) and \(\mathrm{W}'_v\) are the corresponding weight matrices. \(\mathrm{W}'_k\) and \(\mathrm{W}'_v\) are trainable.

Given the output hidden feature from the Meta-Controller \( y_{meta2}\), \(\mathrm{Z}''\) can be computed as:
\begin{equation}
\mathrm{Z}''=\,\,\mathrm{Attention}\left( \mathrm{Q},\mathrm{K}'',\mathrm{V}'' \right) \,\,=\,\,\mathrm{Softmax} \left( \frac{\mathrm{QK}''^{\top}}{\sqrt{\mathrm{d}}} \right) \mathrm{V}''
\end{equation}
where \(\mathrm{Q}=\,\,f_{base}\mathrm{W}_q\), \(\mathrm{K}'' =\,\,y_{meta2}\mathrm{W}''_k\), \(\mathrm{V}'' =\,\,y_{meta2}\mathrm{W}''_v\) are the key, and values matrices from the Meta-Controller, \(\mathrm{W}''_k\) and \(\mathrm{W}''_v\) are the corresponding weight matrices. \(\mathrm{W}''_k\) and \(\mathrm{W}''_v\) are trainable.

Then, we add three attention computation results together and get the final formulation of the decoupled triple attention:
\begin{equation}
\mathrm{Z}^{final}\,\,=\,\,\mathrm{Z}+\mathrm{Z}'+\mathrm{Z}'' 
\end{equation}
where \(\mathrm{Z}^{final}\) is the final attention layer we send back to the U-Net of multi-view diffusion model and replace the original attention layer Z.




This decoupled architecture operates via a three-branch parallel design. Specifically, \(y_{meta1}\) is injected early into the Base-Adapter to selectively gate \(I_{ref}\) features during extraction, while \(y_{meta2}\) provides a direct residual attention map (\(\mathrm{Z}''\)) to correct or supplement the base signal (\(\mathrm{Z}'\)) at the final stage. By leveraging this decoupled triple attention, the model learns higher-level abstract features and effectively prevents negative interference from misaligned references.

\begin{figure*}[!t]
    \centering
    \includegraphics[width=0.95\linewidth]{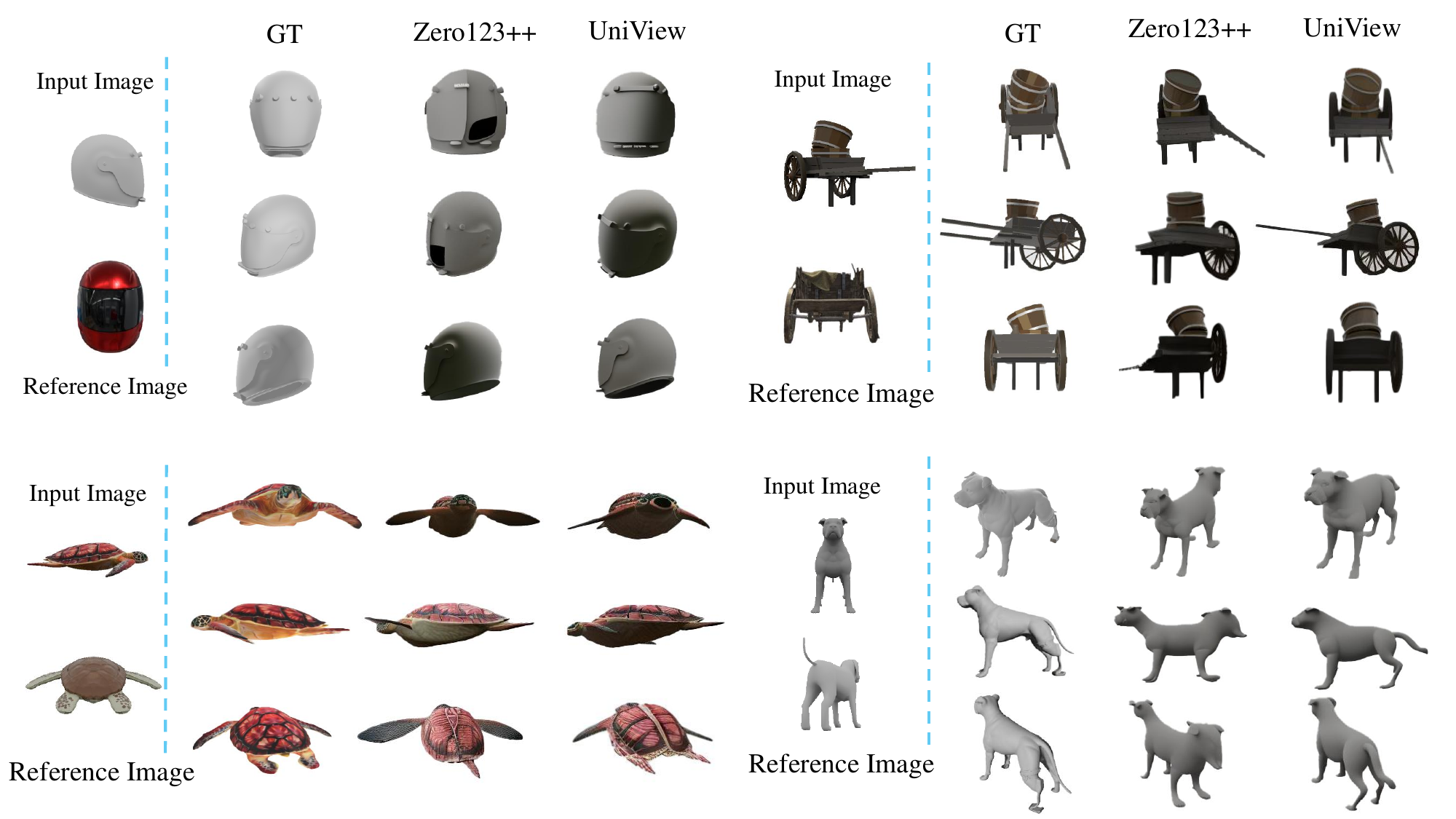}
    \vspace{-0.4cm}
    \caption{Qualitative results of UniView and the baseline.}
    \label{figure4}
    \vspace{-0.5cm}
\end{figure*}

\section{Experiments}
\label{sec:experiments}
In this section, we present the experimental deployment and results of UniView. In Section \ref{Dataset preparation}, we provide detailed descriptions of the dataset preparation. In section \ref{RAG Database construction}, we introduce how we construct our database for the retrieval system. In Section \ref{quantitative and qualitative results}, we introduce the quantitative and qualitative results of UniView on single-image novel view synthesis. In Section \ref{ablation studies}, we present ablation studies. We employ Zero123++ V1.2\cite{zero123++} as the base pre-trained model for our experiments, using clip-vit-large-patch14\cite{clipencoder} as the image encoder and set inference steps to 75.

\subsection{Dataset preparation}
\label{Dataset preparation}
Objaverse\cite{objaverse} is a large dataset which contains over 800K 3D models with descriptive captions, tags, and animations. The Objaverse-LVIS subset comprises 46,000 3D objects spanning 1,156 categories, with all models annotated with LVIS (Large Vocabulary Instance Segmentation)\cite{lvis} labels indicating their respective categories. We employ the Objaverse-LVIS subset to construct our dataset. Utilizing the LVIS labels, we sampled 20K groups of 3D objects, each group containing two objects (a and b) from the same category.

For each group of objects, we sampled 7 viewpoints for object A and 1 viewpoint for object B. To make the input view more challenging, we define the camera coordinates when rendering view 1 of object A as: \(x=0\), \(y=-camera\,\,distance\approx -1.0\), \(z=0\), with FOV = 49.13° and focal length = 49.13 mm. The azimuth angles are set to 0°, 90°, 180°, and 270°, with elevation angle = 0° (i.e., fully frontal view, fully lateral view, or fully posterior view). Following Zero123++ V1.2, we defined the camera coordinates for views 2-7 of object A as: relative to view 1, azimuth angles = [30°, 90°, 150°, 210°, 270°, 330°] and elevation angles = [20°, -10°, 20°, -10°, 20°, -10°]. Views 2-7 serve as ground-truth for novel view synthesis. For object B, the rendering viewpoint is defined relative to object A's view 1 with randomly sampled relative viewpoints within azimuth angles 90°-270° and elevation angles -30°-30°, while keeping other parameters unchanged. This configuration ensures that object B's viewpoint forms complementary perspectives relative to object A's view 1.

Following Zero123++ V1.2, we rendered all 20K pairs of 3D objects using Blender and set the background to pure white. Subsequently, we manually eliminated samples with rendering failures, noticeable distortions, or misclassifications and ultimately selected 15K sets of high-quality rendered data to constitute our training and testing datasets.

\begin{table}[t!]
    \centering
    \caption{Quantitative comparison with baselines on 2D and 3D metrics.}
    \vspace{-0.25cm} 
    \renewcommand{\arraystretch}{1.1} 
    
    \begin{tabular}{l c c c | c} 
    \toprule 
    \textbf{Method} & \textbf{PSNR} $\uparrow$ & \textbf{SSIM} $\uparrow$ & \textbf{LPIPS} $\downarrow$ & \textbf{CD} $\downarrow$ \\
    \midrule 
    Zero123     & 12.77 & 0.691 & 0.287 & 0.076 \\
    LGM         & 14.81 & 0.778 & 0.237 & 0.055 \\
    OpenLRM     & 15.05 & 0.802 & 0.198 & 0.042 \\
    EscherNet   & 13.44 & 0.753 & 0.166 & 0.047 \\
    SV3D        & 15.74 & 0.796 & 0.229 & 0.051 \\
    \midrule
    Zero123++   & 14.22 & 0.753 & 0.256 & 0.044 \\
    \textbf{UniView (Ours)} & \textbf{16.99} & \textbf{0.847} & \textbf{0.162} & \textbf{0.040} \\
    \bottomrule 
    \end{tabular}
    \label{table1}
    \vspace{-0.2cm}
\end{table}


\subsection{RAG database construction}
\label{RAG Database construction}
As described in Section \ref{Dataset preparation}, we utilize a database comprising 20,000 images as the data source for our Dynamic Reference Retrieval System. To construct this database, we sampled 5,000 3D objects from the unused portions (i.e., not allocated to training, testing, or validation sets) of the Objaverse-LVIS subset, following LVIS category labels. These objects span 100 distinct categories, with 50 instances per category. Each object was rendered in Blender from four orthogonal viewpoints, corresponding to azimuth angles of [0°, 90°, 180°, 270°] at a fixed elevation angle of 0°. All other camera parameters and background settings remain consistent with those specified for the dataset in Section \ref{Dataset preparation}.

\begin{table}[t!]
    \centering 
    
    \vspace{-0.2cm} 
    
    \caption{Quantitative results of the impact of reference image quality on model performance.}
    
    \vspace{-0.25cm} 
    
    \renewcommand{\arraystretch}{1.1} 
    
    \begin{tabular}{l c c c}
        \hline
        \textbf{Reference Image} & \textbf{PSNR} $\uparrow$ & \textbf{SSIM} $\uparrow$ & \textbf{LPIPS} $\downarrow$ \\
        \hline
        Same-category & 16.99 & 0.847 & 0.162 \\
        \textbf{Identical} & \textbf{17.32} & \textbf{0.855} & \textbf{0.158} \\
        Irrelevant & 15.76 & 0.661 & 0.243 \\
        \hline
    \end{tabular}
    \label{table2}
    \vspace{-0.6cm}

\end{table}

\subsection{Quantitative and qualitative results}
\label{quantitative and qualitative results}
The qualitative results of UniView are shown in Figure \ref{figure4}. It can be observed that under challenging viewpoint conditions of the input image, the base model (Zero123++) exhibits significant artifacts in synthesized novel views, manifesting phenomena such as helmet visors being only partially rendered or dogs appearing with two heads - clear deviations from ground-truth. When introducing a reference image through our UniView framework, the generation quality demonstrates substantial improvement, with no recurrence of such artifacts.


\begin{table}[t!]
\centering
\vspace{-0.2cm}
\caption{User Study}
\vspace{-0.25cm}
\label{tab:user_study}
\resizebox{0.9\linewidth}{!}{
\begin{tabular}{l|cccccc|c}
\toprule
\textbf{Method} & Zero123 & LGM & OpenLRM & EscherNet & SV3D & Zero123++ & \textbf{UniView} \\
\midrule
\textbf{Score}$\uparrow$ & 2.7/5 & 3.6/5 & 3.9/5 & 3.2/5 & 3.6/5 & 3.3/5 & \textbf{4.1/5} \\
\bottomrule
\end{tabular}
}
\label{table4}
\vspace{-0.6cm}
\end{table}


We conducted a quantitative evaluation on the Objaverse dataset using 100 randomly selected pairs. We compared UniView against several single-image NVS and 3D reconstruction baselines, including Zero123 \cite{zero123}, LGM \cite{LGM}, OpenLRM \cite{openlrm}, EscherNet \cite{eschernet}, SV3D \cite{sv3d}, and our backbone Zero123++ \cite{zero123++}. We first report standard visual metrics (PSNR, SSIM, LPIPS). Additionally, to evaluate 3D consistency, we reconstructed meshes using One-2-3-45++ \cite{One-2-3-45++} and measured the Chamfer Distance (CD). As shown in Table \ref{table1}, UniView outperforms all baselines across both 2D and 3D metrics.



We investigated the impact of reference quality by testing three settings: same-category, identical, and irrelevant objects. As shown in Table \ref{table2}, using identical objects yields only marginal improvements over same-category ones, confirming our method's robustness. Conversely, irrelevant references cause significant degradation. We attribute this to the training distribution: the model is trained on semantically consistent pairs and thus struggles when guided by unrelated visual cues.

We also conducted a user study, inviting 20 participants to rate the results generated by UniView and the baseline models. The results are shown in Table \ref{table4}. Our findings indicate that users demonstrate a clear preference for the outcomes produced by our model.

\subsection{Ablation studies}
\label{ablation studies}

Ablation studies shown in Table \ref{table3} validate Uniview's design. To verify the effectiveness of our Meta-Adapter architecture, we firstly employed the Base-Adapter alone and then employed Base-Adapter with Meta-Controller (without zero convolution isolation). Variants (b) and (c) show that naively injecting reference features or Meta-Controller signals without isolation disrupts pre-trained priors due to initialization interference.To verify our Decoupled Triple Attention, we employed a "Joint Attention" variant (d) by concatenating features into a single cross-attention layer. The performance drop in (d) proves that joint fusion causes feature dilution between reference and control signals. Our full model (e) achieves the best results , confirming that the complete Meta-Adapter architecture and the decoupled triple attention mechanism are essential for isolating the backbone while effectively integrating multi-branch features.

\begin{table}[t!]
    \centering
    \vspace{-0.2cm}
    \caption{Quantitative results of ablation studies.}
    \vspace{-0.25cm}
    \renewcommand{\arraystretch}{1.1}
    \resizebox{0.95\linewidth}{!}{
    \begin{tabular}{l l c c c}
    \hline
    \textbf{Variant} & \textbf{Method} & \textbf{PSNR} $\uparrow$ & \textbf{SSIM} $\uparrow$ & \textbf{LPIPS} $\downarrow$ \\
    \hline
    (a) & Zero123++ & 14.22 & 0.753 & 0.256 \\
    (b) & w/ Base-Adapter & 12.01 & 0.664 & 0.298 \\
    (c) & w/ Meta-Adapter(Non-isolated) & 13.42 & 0.789 & 0.275 \\
    (d) & w/ Joint Attention & 15.35 & 0.802 & 0.221 \\
    (e) & \textbf{UniView(Full)} & \textbf{16.99} & \textbf{0.847} & \textbf{0.162} \\
    \hline
    \end{tabular}}
    \label{table3}
    \vspace{-0.6cm}
\end{table}




\section{Conclusions}
\label{sec:conclusion}

In this paper, we present UniView, a single-image novel view synthesis model enhanced by reference images. Through three mechanisms: the Dynamic Reference Retrieval System, Meta-Adapter and Decoupled Triple Attention, UniView automatically selects reference images and effectively injects reference signals into a pre-trained multi-view diffusion model while mitigating negative interference of reference signals on the original model. UniView significantly improves the quality of single-image novel view synthesis under challenging input views, particularly for completely invisible regions in the input view. Furthermore, UniView provides an enhanced foundational component for downstream novel view synthesis-based tasks such as single-image 3D reconstruction.

\bibliographystyle{IEEEbib}
\bibliography{icme2026references}

\end{document}